%% file: final_version.tex
\documentclass[final]{cvpr}

\usepackage{times}
\usepackage{epsfig}
\usepackage{graphicx}
\usepackage{amsmath}
\usepackage{amssymb}
\usepackage{algorithm}
\usepackage{algpseudocode}
\usepackage{multirow}
\usepackage{booktabs}

\input{vcl-shortcuts.tex}

\usepackage[pagebackref=true,breaklinks=true,colorlinks,bookmarks=false]{hyperref}

\def\cvprPaperID{5680} 
\def\confYear{CVPR 2021}

\begin{document}


\title{TPCN: Temporal Point Cloud Networks for Motion Forecasting}

\author{
	\begin{tabular}{ p{2.8cm}<{\centering} p{2.8cm}<{\centering} p{2.8cm}<{\centering}}
Maosheng Ye\textsuperscript{1$\ddagger$} & Tongyi Cao\textsuperscript{2}  & Qifeng Chen\textsuperscript{1} 
\end{tabular}\\
\textsuperscript{1}Hong Kong University of Science and Technology  \quad  \textsuperscript{2}DEEPROUTE.AI
%
}

\maketitle

\begin{abstract}
   We propose the Temporal Point Cloud Networks (TPCN), a novel and flexible framework with joint spatial and temporal learning for trajectory prediction. Unlike existing approaches that rasterize agents and map information as 2D images or operate in a graph representation, our approach extends ideas from point cloud learning with dynamic temporal learning to capture both spatial and temporal information by splitting trajectory prediction into both spatial and temporal dimensions. In the spatial dimension, agents can be viewed as an unordered point set, and thus it is straightforward to apply point cloud learning techniques to model agents' locations. While the spatial dimension does not take kinematic and motion information into account, we further propose dynamic temporal learning to model agents' motion over time. Experiments on the Argoverse motion forecasting benchmark show that our approach achieves state-of-the-art results.

\end{abstract}

\section{Introduction}

\let\thefootnote\relax\footnotetext{\textsuperscript{$\ddagger$}Part of the work was done during an internship at DEEPROUTE.AI.}

Motion forecasting in autonomous driving concerns future trajectories of agents, including vehicles and pedestrians. For a self-driving car, the predicted future trajectories of surrounding traffic participants serve as key information to plan its future trajectories. A self-driving car should be able to predict the distribution or a few possible future trajectories of each agent as the future is full of uncertainty, given the relevant sensor input information in the past.

Traditional methods for motion forecasting~\cite{houenou2013vehicle, schulz2018interaction,xie2017vehicle,ziegler2014making} are based on kinematic constraints and road map information with handcrafted rules. 
Though these approaches are sufficient in many simple situations, they fail to capture the rich behavior strategies and interaction in complex urban scenarios.
Great progress has been made to explore the power of data-driven methods in motion forecasting with deep learning~\cite{bansal2018chauffeurnet, chai2019multipath,cui2019multimodal, djuric2018short, phan2020covernet}. These methods encode the agents (e.g., vehicles, pedestrians, and cyclists) and high-definition map (HD map) 
information by rasterizing the corresponding elements (lanes, crosswalks) as lines and polygons with different colors. A standard image backbone network~\cite{he2016deep, simonyan2014very} is then applied to the rasterized image to extract the map and agent features and perform motion prediction. 

\begin{figure}
   \centering
   \includegraphics[width=\linewidth]{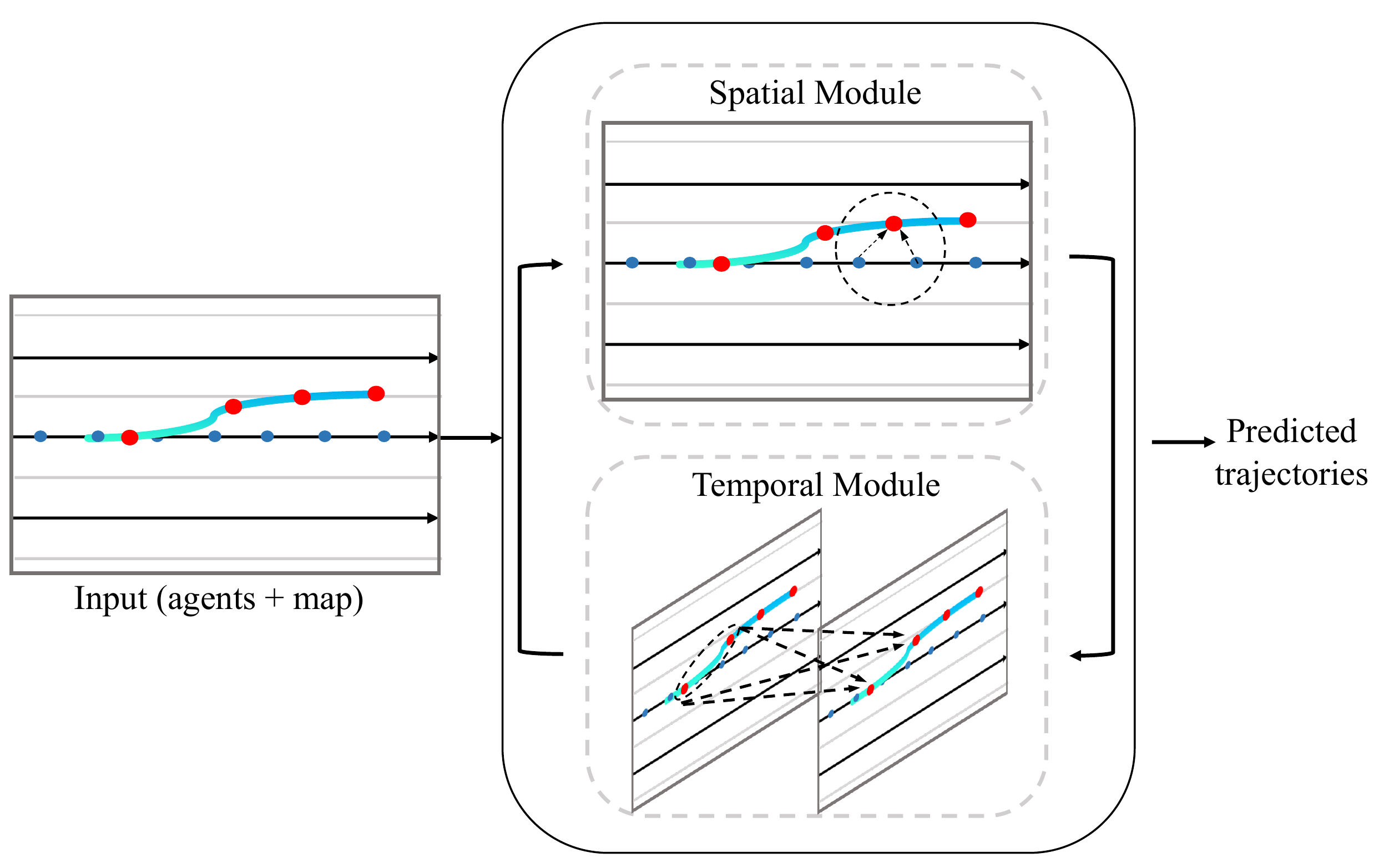}
   \caption{A high-level illustration of our approach. Red points represent the agent history trajectory points, while blue points are discrete map lane points. We use a spatial module based on point cloud learning to extract geometric features and a temporal module to extract sequential features. Both modules utilize the output of the other module and propagate mutually to output future trajectory points.}
   \label{fig:overall_pipe}
\end{figure}
However, the rasterized image is an overly complex representation for environment and agent history and requires significantly more computation and data to train and deploy.
More succinct representations have been explored to avoid this heavy process.
VectorNet~\cite{gao2020vectornet} 
proposes a vector representation to exploit the spatial locality of individual road components with graph neural networks. LaneConv~\cite{liang2020learning} constructs a lane graph from vectorized map data and proposes LaneGCN to capture the topology and long dependency of the agents and map information. 
Both VectorNet~\cite{gao2020vectornet} and LaneConv~\cite{liang2020learning} can be viewed as extensions of graph 
neural network in prediction with strong capability to extract spatial locality. 
However, both works fail to fully utilize the temporal information of agents with less focus on temporal feature extraction. 

In this work, we extend ideas from 3D point cloud learning to the motion forecasting task. Previous work on point cloud network focuses on spatial points. We extend the metric space to the joint spatial-temporal space and represent the agents' history observations and map data as points in this space.
Since the raw input data of prediction is a set of points that contain different agents with historical observations and map data, spatial and temporal learning will be two key components in prediction learning. 
Ignoring either information will lead to information loss and reduce the model's capability of context learning.

In order to combine spatial and temporal learning in a flexible and unified framework, we propose Temporal Point Cloud Networks (TPCN). Compared with GCN based methods~\cite{gao2020vectornet,liang2020learning}, our TPCN does not manually specify the interaction structures (e.g., connectivity in the graph) and avoid the complex correlation learning process.
TPCN models the prediction learning task as joint learning between a spatial module and a temporal module. In the spatial module, note that the waypoints and map points have very similar properties as point clouds, both being sparse and permutation invariant, and have a strong geometric correlation. Thus, point cloud learning strategies can be effective for spatial feature extractions. 
Instead of directly applying works~\cite{qi2017pointnet, qiPointNetDeepHierarchical2017a,thomas2019kpconv, ye2020hvnet} whose computation cost is high, we propose our novel spatial learning layer, namely \textbf{Dual-representation Spatial Learning} to obtain pointwise and voxelwise features through point cloud learning. 
Meanwhile, we propose \textbf{Dynamic Temporal Learning} in the temporal module to effectively extract the time-series information and motion estimation. Compared with traditional \textit{Hard Temporal Learning}~\cite{bai1803empirical, lea2017temporal,liang2020learning}, 
our dynamic temporal learning layer naturally handles variant time length of different agents in the same sequence without the need to pad the history.
By switching between the two modules, the spatial features and temporal features from these two modules are propagated mutually, each module taking the features of the other module as input. As such, spatial learning will utilize the temporal information (e.g., motion state), while the temporal learning will have some spatial guidance (e.g., map information), namely \textbf{Joint Learning}. Fig.~\ref{fig:overall_pipe} illustrates the overall architecture of our approach. Note that we model the selection of multi-modal trajectories problem as displacement regression rather than classification.  

In summary, our contributions reside as follows:
\begin{itemize}
   \item We propose a novel and flexible architecture for prediction learning, which models the complex process as joint spatial and temporal learning. Dual-representation Spatial Learning for feature extraction of waypoints and map data is proposed as the spatial module. Meanwhile, we propose novel Dynamic Temporal Learning, consisting of Multi-interval Learning and Instance Pooling. 
   \item We propose displacement prediction for multi-modal trajectories selection, which alleviates the hard assignment in classification through regression.
   \item Extensive experiments are conducted on the large-scale Argoverse motion forecasting benchmark to show the effectiveness of our approach.
\end{itemize}

\section{Related Work}
Most existing works on prediction can be roughly divided into three categories according to their representation and architecture.

\noindent\textbf{Rasterization based methods}. Rasterization BEV images are the most common and direct ways to represent the structure of map and neighborhood relationships among agents. 
Some methods~\cite{bansal2018chauffeurnet,casas2018intentnet,luoFastFuriousReal2018,phan2020covernet} render the HD map elements (junctions, lanes) as BEV images with different colors according to their types. In the format of images, a series of standard convolution layers or backbones~\cite{he2016deep} can be applied to simplify the prediction task as trajectories selection and offset regression problems. Furtherly, some
works~\cite{chai2019multipath,phan2020covernet} propose to use anchor trajectories with human prior knowledge based on motion constraint to make the results more consistent with the current dynamic state and alleviate the difficulties in multi-modal prediction. However, these approaches have internal limitations since the performance is highly related to the spatial resolutions of the rasterized images. The temporal information can not be represented or modeled in the rendered images intuitively. 

\noindent\textbf{GCN based methods}. Graph Convolutional Network (GCN)~\cite{duvenaud2015convolutional,henaff2015deep,shuman2013emerging} nowadays gains its popularity in processing non-structural data and dealing with correlation relationship. Compared with traditional CNN, GCN shows its great promise in capturing the spatial locality on both euclidean and non-euclidean structure. 
With adjacency matrices, GCN focuses on learning the relationship between graph nodes and vertices. VectorNet~\cite{gao2020vectornet} introduces a novel vector representation and applies a graph neural network to predict the intent of vehicles. M Liang \textit{et al.}~\cite{liang2020learning} proposes LaneGCN based on GCN, which is a specialized version designed for lane graphs. In order to capture the complex topology of HD map effectively, 
it combines with multi-scale dilated convolution. Social-STGCNN~\cite{mohamed2020social} models interactions as a graph by defining a novel kernel function to learn spatial and temporal patterns from pedestrian trajectories and behavior. However, GCN based methods are often faced with an efficiency problem when dealing with large scale scenarios which contain lots of nodes and vertices.  

\noindent\textbf{Hybird Methods}. In order to provide more interpretable and kinematic constraints, some works~\cite{casas2018intentnet, mangalam2020not, zhao2020tnt, zeng2019end} decouple the task as two-stage. Firstly, they discretize search space via uniform sampling or based on HD map to generate some proposals. Compared with anchor trajectories, these proposals can be more stable and more informative to capture the uncertainty of the multi-modal prediction. 
Secondly, they encode the HD map and agents via vector representation or rasterized images to furtherly refine each proposal. To some extent, two-stage methods can fully incorporate expert knowledge or classical planning or prediction approaches. On the other hand, it means the final outputs of refinement networks have a strong dependency on the quality of the proposals, which requires a reasonable sampling strategy or mature planning module. 

In comparison with these methods above, our TPCN has a novel representation and architecture, including spatial learning based on dual-representation point cloud learning and dynamic temporal information learning. We split the task into submodules to capture both spatial and temporal information effectively.

\begin{figure}[t]
    \centering
    \includegraphics[width=7.5cm]{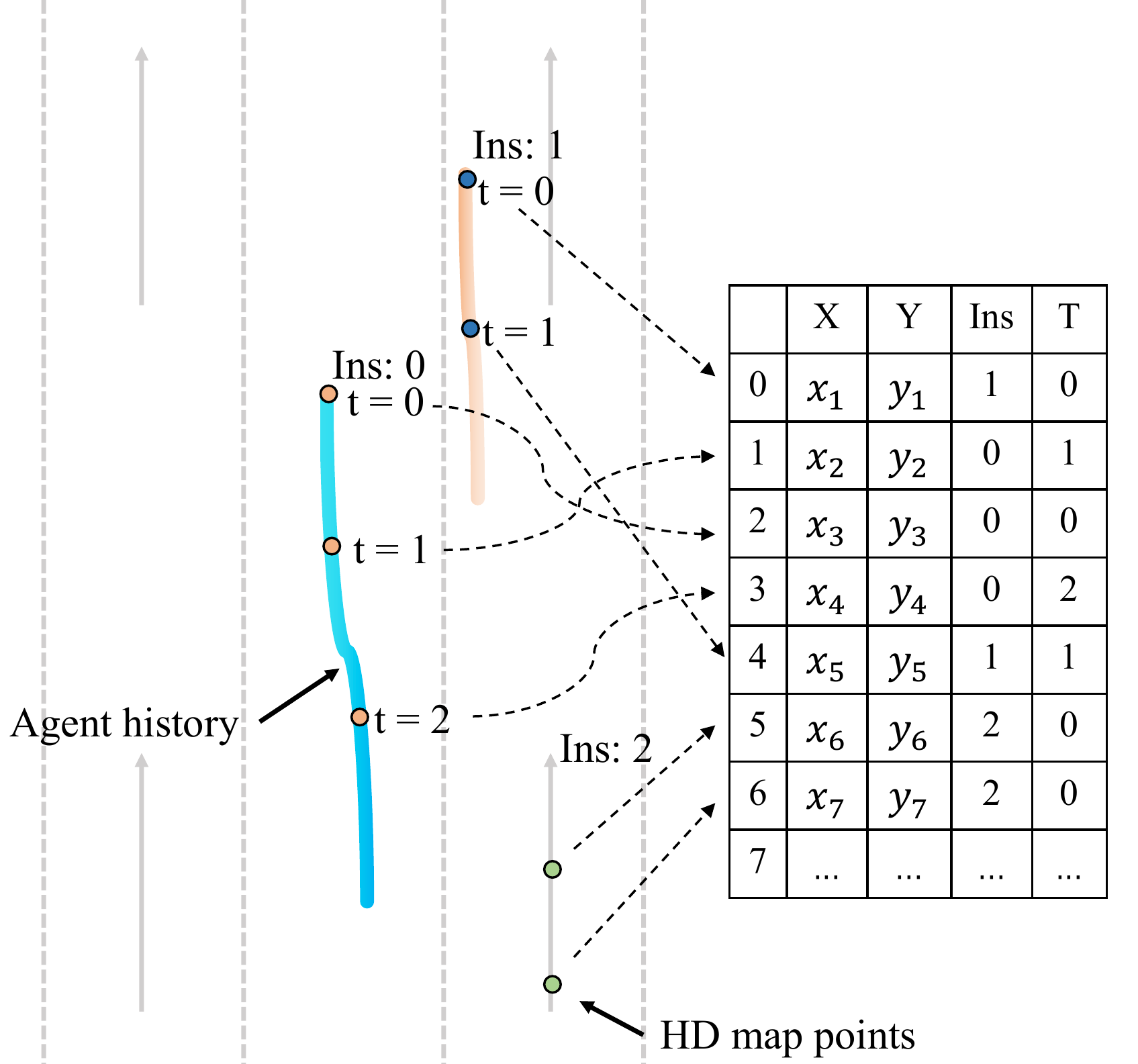}
    \caption{An example that shows the instance time indexing system. The points with the same color belong to the same instance.}
    \label{instance-time-indexing system}
    \vspace{-5px}
\end{figure}

\section{Approach}

The overall network architecture of our TPCN approach consists of two modules: 1) Dual-representation Spatial Learning and 2) Dynamic Temporal Learning. The \textbf{Dual-representation Spatial Learning} serves as the spatial module to model spatial features and the \textbf{Dynamic Temporal Learning} is the temporal module to extract temporal features. Both modules are integrated to propagate features mutually in spatial and temporal dimensions to achieve \textbf{Joint Learning}. 
As shown in Fig.~\ref{fig:overall_pipe}, each module takes the pointwise features of the other module as input to generate corresponding pointwise output features, which is a natural foundation for fusing pointwise context information across multiple domains. 


Typically, the motion prediction task will contain agents' data and environment information encoded with map data. 
We define an agent instance as an agent with a set of trajectory points. 
While map data refers to static objects without temporal information, a map instance can be described as an ordered point set for one specific element\ (e.g., a piece of lane centerline points).
Thus, agent data will be represented by 
$\left\{ {{{\bf{p}}_{i, 1}}}, {{{\bf{p}}_{i, 2}}}, \hdots,  {{{\bf{p}}_{i, T_i}}} \right\}$, where ${{\bf{p}}_{i, t}}$ 
means the $i$-th agent's coordinate at time \textit{t}, and $T_i$ is the time sequence length for $i$-th agent. 
Meanwhile, we represent map data in the format of $\left\{ {{{\bf{p}}_{i, 1}}}, {{{\bf{p}}_{i, 2}}}, \hdots,  {{{\bf{p}}_{i, N_i}}} \right\}$, where 
${{\bf{p}}_{i, j}}$ is the $j$-th point of $i$-th map element instance with $N_i$ points in total.

\textbf{Voxelization}. Given a grid size $s$, we can construct the mapping from a point ${{\bf{p}}_{i}}=(x_i, y_i)$\ to its voxel index ${{\bf{v}}_{i}}$:
\begin{equation}
   {{\bf{v}}_{i}} = \left( \lfloor{x_i/s}\rfloor, \lfloor{y_i/s}\rfloor\right),
\end{equation}
where $\lfloor{\cdot}\rfloor$ is a floor function. Thus we can build a hash table for the conversion between point coordinate space and voxel coordinate space $\left\{{\bf{p}}_{i}, {\bf{v}}_{i}\right\}$. 

\textbf{Instance Time Indexing}. Apart from voxel and point spaces, we formulate the temporal space indexing system to address the dynamic and different sequential lengths of different agents. 
We represent all the instances over time as $\{\bf{m}_i\}$, where the $i$-th element ${{\bf{m}}_{i}}$ = $\left(ins_{i}, t_{i}\right)$ is an instance time index referring to the $t_{i}$-th trajectory point of instance $ins_{i}$.  See Fig.~\ref{instance-time-indexing system} for an example. A special case is that $t_{i}$ is zero for all static instances in the map data.

Both voxelization and instance time indexing aim to provide space mapping or hashing: voxelization maps from the Cartesian coordinate system to a structural grid representation and instance time indexing maps from an index to a trajectory point of an instance. These mapping or hashing systems provide convenience for feature transformation between different spaces or representations.




\subsection{Dual-representation Spatial Learning}
\label{sec:dual-representation spatial}
\begin{figure}[t]
    \centering
    \includegraphics[width=\linewidth]{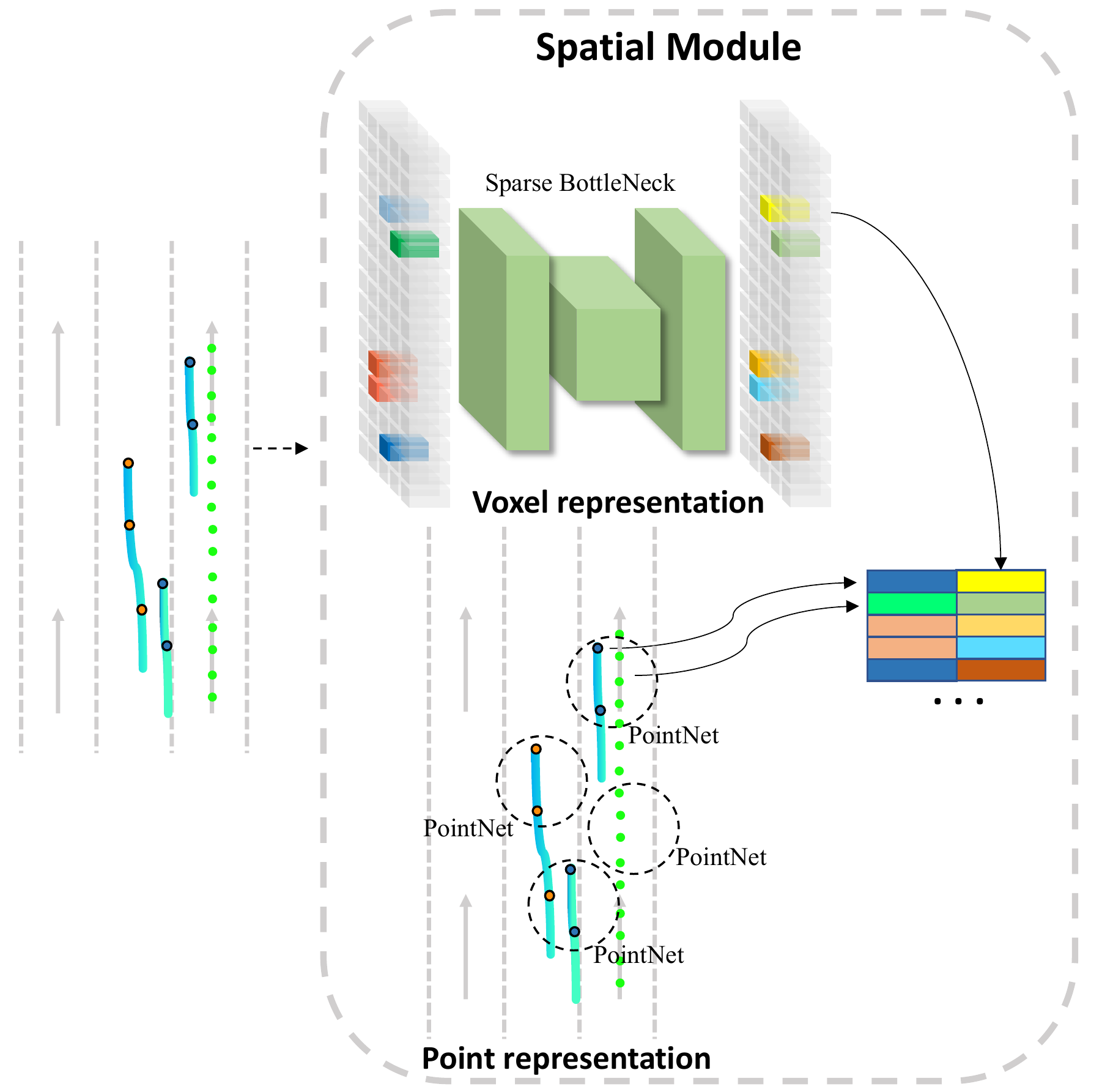}
    \caption{Dual-representation Spatial Learning.}
    \label{spatial module}
\end{figure}
For the spatial module, we choose a point cloud learning approach to retrieve the spatial features of waypoints and map data with their locality and spatial geometry.  
Inspired by recent multiple representations learning~\cite{shi2020pv, tang2020searching} that shows advantages over a single representation~\cite{qiPointNetDeepHierarchical2017a, thomas2019kpconv}, 
we propose a dual-representation method to leverage the merits of mutually complementary information between voxel and point representations. The overall architecture of the spatial module is illustrated in Fig.~\ref{spatial module}. 

\textbf{Pointwise Feature Learning}. Pointwise features maintain geometric information and neighborhood relationship for interactions among points. 
To this end, we utilize the hierarchical feature learning from PointNet++~\cite{qiPointNetDeepHierarchical2017a} for pointwise feature extraction at different levels of the local neighborhood to exploit more local structure and correlation. 

\textbf{Point-Voxel Feature Propagation}. Inspired by HVNet~\cite{ye2020hvnet}, we can transform pointwise features to voxel space by scattering operations. In this process, 
we maintain the hash table for each point, which stores the key and value pairs to map a Cartesian coordinate to a voxel index. 
Then, following the Feature Transformation Propagation algorithm (FTP)~\ref{algo_p_v}, we put all the points that share the same voxel index into the same cluster (each voxel may contain more than one point). Finally, we calculate the mean features over the points in each cluster as final features for the corresponding voxel, which is similar to average pooling.

\begin{algorithm}[h]
   \caption{Feature Transformation Propagation}
   \begin{algorithmic}[1]
     \Require
     All pointwise features $PF$ with corresponding indexing hash table $H$
     \State $\text{clusters}$ = \{\}
     \For{each $PF_i\in PF$}
      \State $\text{clusters}.\text{at}(H_{i}).\text{append}(PF_i)$
     \EndFor
     \For{each $H_{i} \in H$}
      \State $\text{clusters}.\text{at}(H_{i})$ = $\text{mean}(\text{clusters}.\text{at}(H_{i}))$
     \EndFor
 
     \State\Return $\text{clusters}$
   \end{algorithmic}
   \label{algo_p_v}
 \end{algorithm}

\textbf{Voxelwise Feature Learning}. Voxelwise features have a strong capability to extract semantic context information~\cite{liu2019point}.
Most existing works apply 2D or 3D CNN to the rasterized or voxelized images for feature extraction in order to exploit the structural data with the current popular backbone~\cite{he2016deep}.
One of the key parameters for these methods is the grid size. Smaller grid size leads to less information loss but brings higher computation cost and latency. Meanwhile, compact 2D or 3D tensor of rasterized images neglect the sparsity of the input data and involve plenty of non-activated regions that may mess up feature learning.  

Therefore, we employ the sparse convolution~\cite{graham20183d, yan2018second} as our feature extractor to afford a smaller grid size for fine-grained voxelwise features.  
Furthermore, we build a Sparse BottleNeck network with skip connections, which replaces the bottleneck blocks with sparse convolutions in ResNet~\cite{he2016deep}. 
Stacking Sparse BottleNeck layers not only quickly expands the receptive field at a low computational cost but also keeps the activation sparse. 

\textbf{Voxel-Point Feature Propagation}. With voxelwise features, feature propagation from voxel representation to point representation can be performed by the naive nearest neighbor interpolation. 
PVCNN~\cite{liu2019point} interpolates the pointwise features with corresponding neighboring voxelwise features. Since the interpolation weights are based on the physical distance to the neighboring grids accordingly, we extend the weights to be learnable by applying MLP to distance embedding that also concatenates associated voxelwise features. 

\textbf{Dual-representation Fusion}. With pointwise and voxelwise features, we fuse these two types of features by feature concatenation. Thus, we obtain the features with dual representations and higher context information, which will be passed to the next stage of dynamic temporal learning. 
\subsection{Dynamic Temporal Learning}
\label{sec:Dynamic Temporal Learning}
In a motion prediction task, different agents have different lengths of observed past trajectories due to the different lifespan of each agent. Existing methods~\cite{hochreiter1997long, liang2020learning} pad the agents' data whose size is smaller than a given size $T$ 
with zero in order to process data with the same length. We name this operation as \textit{Hard Temporal Learning} (HTL).
HTL has two main drawbacks: 1) padding data will introduce extra unnecessary computation cost, especially when the agents only appear in very few shots; 2) processing padded data will lead to the feature confusion problem, especially when invalid padding data involves the feature propagation. HTL forces the network to capture useless information. 

We propose Dynamic Temporal Learning to address these limitations. Instead of padding, we only preserve the originally provided information without the requirement for a fixed time buffer size for each agent data. 
Therefore, we can retain each agent with a dynamic time sequence length. Furthermore, the Dynamic Temporal Learning consists of the following two parts.

\begin{figure}
    \centering
    \includegraphics[width=8cm]{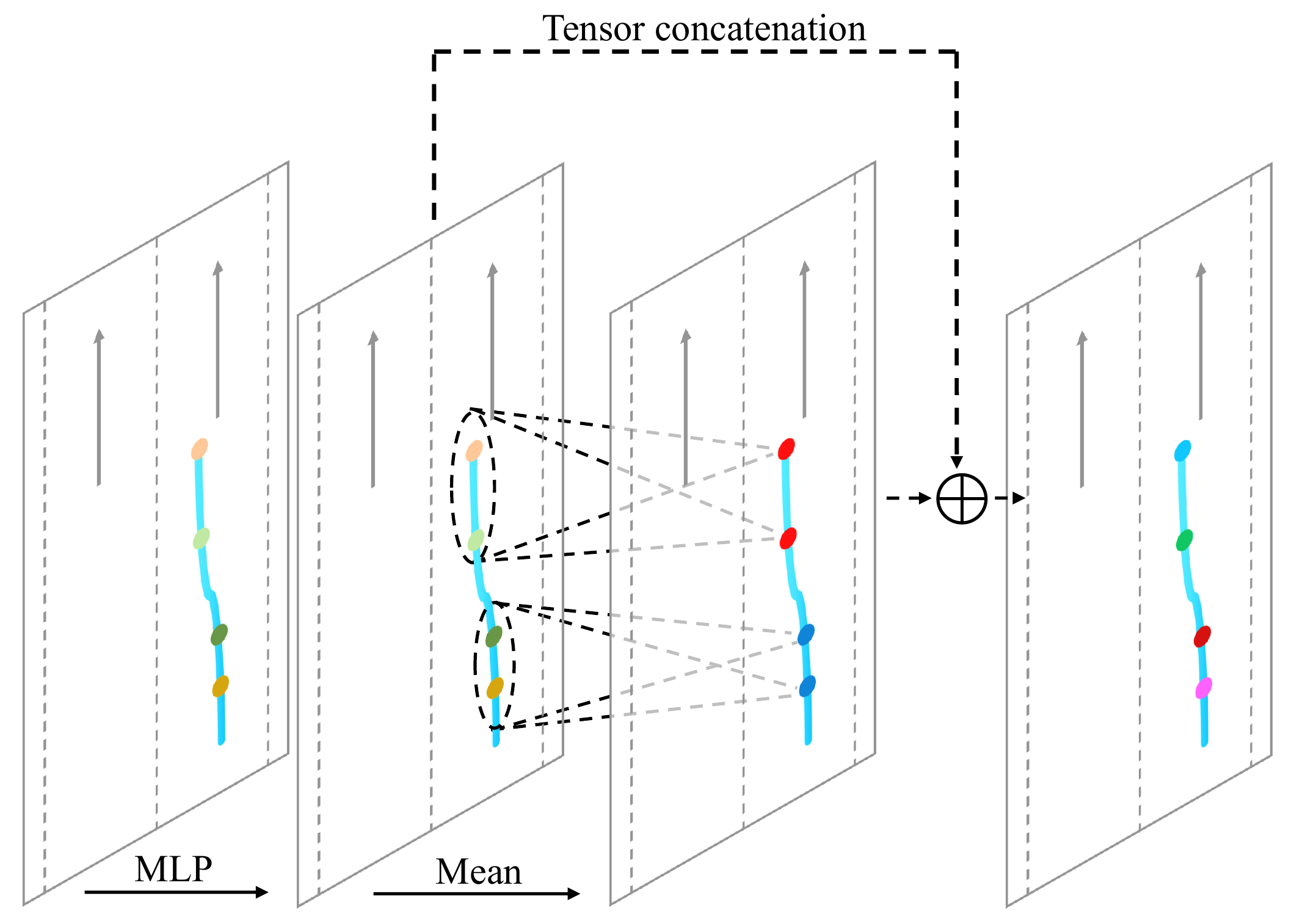}
    \caption{An example of multi-interval learning with an interval size of 2. $\oplus$ means tensor concatenation and points stand for the sequential historical waypoint data for one specific instance.}
    \label{interval_learning_case}
\end{figure}

\textbf{Multi-interval Learning}. The time interval is a key factor for time series data feature learning since it determines the time window size, similar to the receptive field in a CNN. Inspired by the multi-scale or multi-resolution hierarchy that has been proven its effectiveness in capturing local correlations and context, we also exploit Multi-interval Learning with prediction data. However, the main challenge for our application lies in the dynamic property. As a result, we utilize the Instance Time Indexing system and previously introduced FTP~\ref{algo_p_v}, and then perform high efficient Multi-interval Learning (MIL) according to the following Multi-interval Learning algorithm~\ref{algo_mt}.
Given a set of different intervals, we first regroup the instance time indexing to ensure that each point with the same instance ID and timestamp within the same interval will be clustered into the same group. 
Consequently, we employ the FTP algorithm~\ref{algo_p_v} to obtain the mean features of the corresponding interval and map to pointwise features by the following two steps; 1) slicing by using the inverse hash or indexing table to gather the pointwise features and 2) concatenating the input tensor as a shortcut connection. Note that the output feature of the current time interval will be passed to the next level as input to progressively and aggressively capture features at increasingly larger intervals over a multi-resolution hierarchy. Fig.~\ref{interval_learning_case} illustrates a special case for Multi-interval Learning with an interval size of 2. 

\begin{algorithm}[h]
   \caption{Multi-interval Learning}
   \begin{algorithmic}[1]
     \Require
     All pointwise features $PF$ with instance time indexing \textbf{m} and the predefined set of time intervals T
     \Ensure
    All pointwise output features $O_p$
     \State $O_p = PF$
     \For{each $t\in \text{T}$}
      \State $m^{*} = \left({{m[0], \lfloor{m[1]/t}\rfloor}}\right)$
      \State $F_{t} = \text{MLP}(O_{p})$
      \State $O = \text{FTP}(F_{t}, m^{*})$
      \State $O_{p} = \text{slice}(O, m^{*})$
      \State $O_{p} = \text{concat}(O_p, F_{t})$
     \EndFor 
     \State\Return $O_p$
   \end{algorithmic}
   \label{algo_mt}
 \end{algorithm}

\textbf{Instance Pooling}. Instance-level features that aggregate information of an instance over time is a crucial component beyond the point, voxel, and time level features, as the fundamental data in this task is composed of instances as mentioned. 
Moreover, instance-level features can capture the long-range or large time interval dependency under some scenarios. For example, when the start point and endpoint of a lane centerline are  far away from each other, it is hard to design a suitable architecture to handle this dependency or correlation. To this end, we propose our Instance Pooling (Ins-Pool) to provide a more flexible way for the instance feature extraction. 
In this process, the Instance Pooling can be viewed as a special case of Multi-interval Learning as it applies pooling operation along with the set of points with the same instance ID to extract global entity features.      

For implementations of the above algorithm~\ref{algo_p_v} and ~\ref{algo_mt}, we optimize and utilize GPU-based scatter, gather and hash table operations to increase runtime efficiency by parallel. 

\subsection{Displacement Prediction and Learning}
A prediction header is built to predict the final forecasting that takes the features from fusion features of spatial and temporal modules as input. Most existing works~\cite{chai2019multipath, cui2019multimodal, liang2020learning} predict K possible trajectories with their confidence scores respectively to model the multi-modal property. The loss is often split into regression and classification parts. During training, the loss will only be backpropagated at the trajectory that has the minimum displacement error at the endpoint. However, loss of classification for trajectory selection based on positive and negative samples assignment is not reasonable and too handcrafted, especially when two trajectories have very close displacement error.
Inspired by the popular concept of IoU prediction~\cite{jiang2018acquisition} in object detection, we predict the final displacement error rather than the trajectory's confidence. Therefore, we alleviate the problem of classification that requires hard assignments by displacement regression. We define the output of our prediction header as $\tau_{disp}$ and $\tau_{reg}$:
\begin{align}
   \tau_{disp} &= \{d_{0}, d_{1}, \dots, d_{K-1}\},\\
   \tau_{reg}^{k} &= \{(x_{1}^{k}, y_{1}^{k}), (x_{2}^{k}, y_{2}^{k}), \dots, (x_{T}^{k}, y_{T}^{k})\},
\end{align}
where $\tau_{reg}^{k}$ is the $k$-th predicted trajectory among K possible trajectories, which contains $T$ waypoints with 2D $(x, y)$ vector representation. $\tau_{disp}$ is the predicted displacement error at endpoint associated with each possible trajectory.

\textbf{Loss Functions}. We sum the trajectory regression $\mathcal{L}_{reg}$ and trajectory displacement $\mathcal{L}_{disp}$ as final loss for training.

For trajectory regression, we follow the previous routine that we choose the best trajectory $k^{*}$ whose displacement error with ground-truth trajectory is minimum:
\begin{equation}
   \mathcal{L}_{reg} = \frac{1}{T}\sum_{i=1}^{T}\rho(x_{i}^{k^{*}}, x_{i}^{gt}) + \rho(y_{i}^{k^{*}}, y_{i}^{gt}),
\end{equation}
where $(x_{i}^{gt}, y_{i}^{gt})$ represents the ground-truth coordinate at timestamp $i$, and $\rho$ is a smooth $L_1$ loss function:
\begin{equation}
   \mathcal{L}_{disp} = \frac{1}{K}\sum_{i=1}^{K}\rho(d_{i}, d_{i}^{*}),
\end{equation}
where $d_{i}^{*}$ is the ground-truth displacement between the $k$-th trajectory and ground-truth trajectory.  For inference, we sort the trajectories according to their predicted displacement in an ascending order.

\section{Experiments}
In this section, we conduct extensive experiments on Argoverse~\cite{chang2019argoverse}, which is one of the largest public motion forecasting datasets with rich HD map information. We also evaluate the proposed modules with ablation studies to show their effectiveness. 
\subsection{Experimental setup}
\textbf{Dataset}. Argoverse~\cite{chang2019argoverse} is a public motion forecasting dataset. It has more than 300K 5-second sequences collected in Pittsburgh and Miami. For each sequence, the sampling rate is 10$Hz$, meaning that the interval of the same object appears in the next timestamp is about 0.1s. There are multiple objects with centroid coordinate of time series trajectories within one sequence, with each object tagged as one of the three types, agent, AV, and others. Moreover, each sequence has only one object, tagged with type agent which is required to be predicted the next 3 seconds future horizon in this challenge. We name this agent as the target agent and other vehicles, including AV as other agents similar to ~\cite{gao2020vectornet}. The whole sequences can be split into training, validation, and test set, with 205942, 39472, and 78143 sequences, respectively. The training and validation sets provide the full 5 seconds trajectories for each target agent data. For the test set, only the first 2 seconds trajectories are given. In addition to trajectories data, 
we could query the map data represented by lane centerlines points via a given location and city name.

\textbf{Metrics}. Following the previous works, we also adopt the widely used metrics Average Displacement Error (ADE) and Final Displacement Error (FDE) as criteria. 
ADE is the average displacement error with ground-truth labels over the entire time steps, and FDE is defined as the displacement error at the endpoint. 
For multi-modal prediction evaluation on Argoverse, minADE, and minFDE are also used since it allows multiple forecasted trajectories. During the evaluation, it selects K trajectories and computes minimum ADE and minimum FDE as metrics. Miss Rate(MR) is also considered in this task, which is the percentage of the predicted trajectories within a certain threshold ($2m$) of ground truth according to endpoint error.
We take minADE, minFDE, MR for K=1 and K=6 as evaluation metrics in our experiments.

\begin{table}[htbp]
    \small
    \centering
    \def\arraystretch{1.2}
    \begin{tabular}{c|c c c c}
    \hline
    \multirow{2} * {Data split} & \multicolumn{4}{c }{Speed distribution(\%)} \\
    \cline{2-5}
    & $\left[ 0, 5\right)$ & $\left[ 5, 10\right)$ & $\left[ 10, 15\right)$ & $\geq 15$ \\
    \hline
    Train & 26.4 & 42.4 & 26.7 & 4.5 \\
    \hline
    Val & 18.6 & 38.7 & 29.8 & 12.9 \\
    \hline
    Test & 33.4 & 52.6 & 13.1 & 0.9 \\
    \hline
    \end{tabular}
    \caption{Speed distribution on different data splits.}
    \label{data_split}
    \vspace{-5px}
\end{table}
\textbf{Data Augmentation}. As shown in Tab.~\ref{data_split}, the speed distribution of target agents varies on different data splits. The validation set has a larger proportion of high-speed agents, while the average speed of agents on the test set is lower. Therefore, we apply global random scaling with the scaling ratio between $\left[ 0.8, 1.25\right]$. Global random scaling can simulate agents' dynamics at different speeds, which can improve the model's generalization ability. Besides that, we apply randomly point dropout with probability $0.9$ and points location perturbation under normal distribution with mean $0$ and standard deviation $0.2$. 

\textbf{Experiment details}. We apply some similar standard preprocessing steps as previous works~\cite{liang2020learning, fang2020tpnet}. First, we translate all the point data to be centered by the coordinate of agent data at $t=0$. We use the orientation between agent location at $t=0$ and its previous location as the positive x-axis. Then, we set the range starting from $[-48, -48]$ to $[48, 48]$ and filter the points outside the region. For voxelization, we set the grid size $g$ to 0.2m to keep the balance between efficiency and performance. The intervals for Multi-interval Learning are predefined as $[2, 4, 6, 8, 16]$ that is enough for capturing sequential information. TPCN is trained for 36 epochs using a batch size of 32 with Adam~\cite{kingma2014adam} optimizer with an initial learning rate of 0.001. Besides that, the learning rate decays at every 10 epochs in a ratio of 0.1. 

\subsection{Ablation study}
\begin{figure*}[t]
    \centering
    \vspace{-2mm}
    \includegraphics[width=1\textwidth]{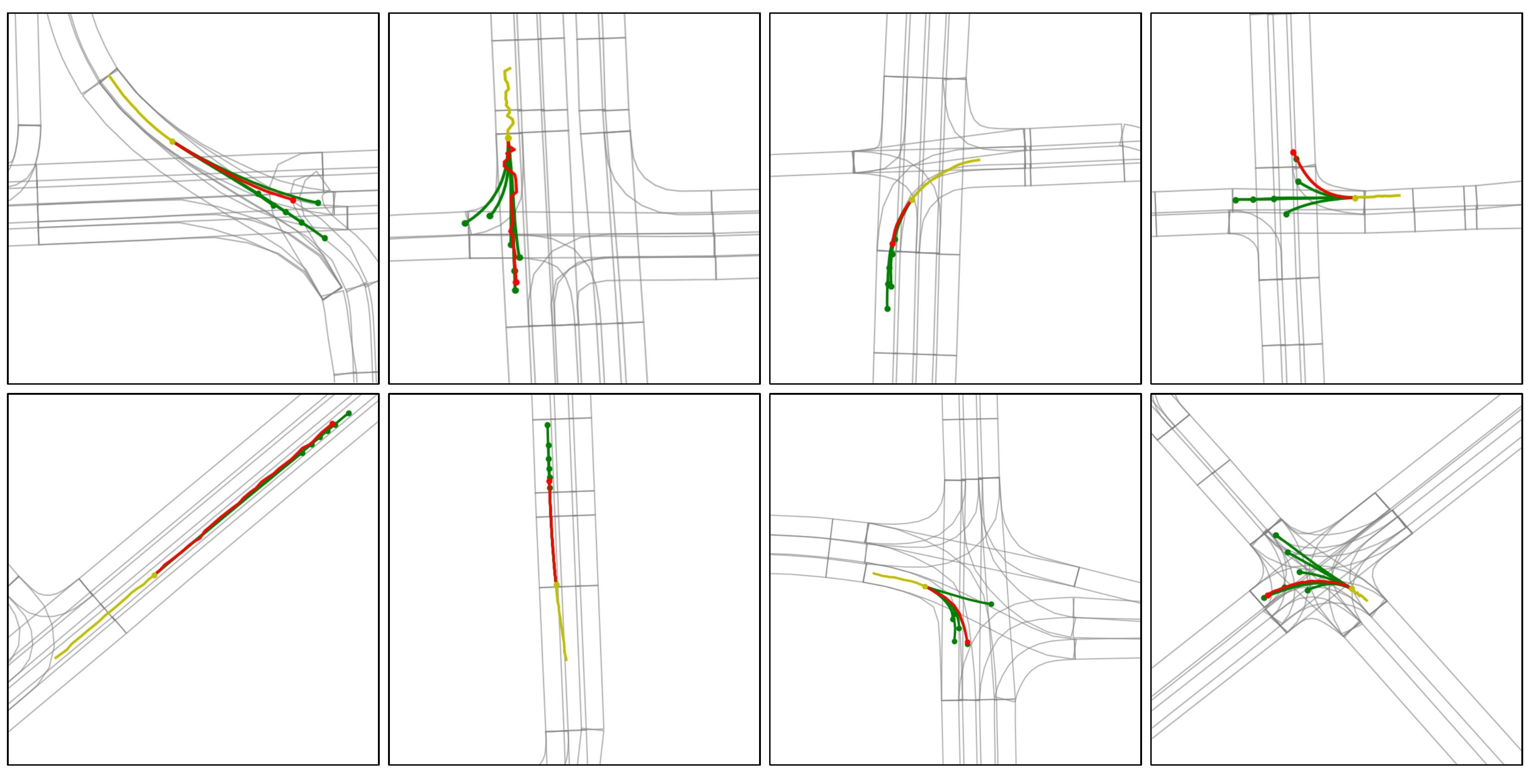}
    \caption{The motion forecasting results on the Argoverse validation set. The past trajectory of the target agent is in yellow, predicted trajectory in green and ground truth in red, respectively. The figures demonstrate the effectiveness of our TPCN on scenarios including left-turn, right-turn, lane change, and so on.}
    \label{fig:prediction_results}
    \vspace{-5px}
\end{figure*}
\begin{table*}[!t]
   \begin{center}
      \small
      \centering
      \def\arraystretch{1.2}
      \begin{tabular}{c c|c c|c|c c c|c c c}
         \hline
         \multicolumn{2}{c|}{ Spatial} & \multicolumn{2}{c|}{ Temporal} &
         \multirow{2}* {Aug}&
         \multirow{2}*{ minADE$_{1}$} & \multirow{2}*{ minFDE$_{1}$} & \multirow{2}*{ MR$_{1}$} & \multirow{2}*{ minADE$_{6}$} & \multirow{2}*{ minFDE$_{6}$} & \multirow{2}*{ MR$_{6}$} \\
         \cline{1-3} 
         \cline{3-4} Point & Voxel & MIL & Ins-Pool & & & & & & \\
         \hline
         \checkmark &  &  &  & & 1.75& 4.00& 0.66&1.01 & 1.88 & 0.28 \\
         \hline
         & \checkmark &  &  & & 1.63& 3.65& 0.62&0.94 & 1.70 & 0.23 \\
         \hline
         \checkmark & \checkmark &  & &  &1.50 &3.31 & 0.57& 0.85&1.42 & 0.16 \\
         \hline
         \checkmark & \checkmark & \checkmark & &&  1.38 & 3.02& 0.52& 0.76& 1.19& 0.13\\ 
         \hline
         \checkmark &  \checkmark & \checkmark & \checkmark & &  1.36&  2.98& 0.51 & 0.74 &  1.18 &  0.12 \\
         \hline
         \checkmark &  \checkmark & \checkmark & \checkmark & \checkmark & \bf 1.34& \bf 2.95& \bf 0.50 &\bf 0.73 & \bf 1.15 & \bf 0.11 \\
         \hline
      \end{tabular}
   \end{center}
   \caption{Ablation study of each component on the Argoverse validation set. Point and Voxel represent pointwise feature learning and voxelwise feature learning, respectively. The temporal module includes Multi-interval Learning (MIL) and Instance Pooling (Ins-Pool). ``Aug'' refers to data augmentation.}
   \label{ablation_study_result}
   \vspace{-5px}
\end{table*}
\textbf{Component study}. We conduct an ablation study on the Argoverse validation set to evaluate and analyze the contributions of our proposed components to the final performance. We take the spatial module without dual representations as our baseline. And then, we add other components gradually, as shown in Tab.~\ref{ablation_study_result}. According to the results, we can draw some conclusions. First, spatial feature extraction based on point cloud learning that takes the unordered set of agents and map points is proven to be effective. With the only spatial module, our model has achieved a strong baseline compared with state-of-the-art methods~\cite{chai2019multipath, gao2020vectornet} that have provided results on the validation set. Second, spatial and temporal modules are both indispensable parts of our model. Spatial module models the geometric information, neighborhood relationship, and interaction among points. The temporal module handles the time series features and focuses more on single instance feature learning. With both modules on, features propagate between spatial and temporal dimensions, thus we achieve the best performance on the validation set. 
Third, point and voxel features are important feature compensation between each other due to the observation that when we apply dual-representation learning, the performance greatly outperforms single representation. Multi-interval learning plays a crucial role in temporal learning to capture sequential information, with about 10$\%$ improvement on displacement metrics. Furthermore, Instance Pooling retrieves the global instance features that address the long-range dependency problem, which leads to better performance. 

\textbf{Displacement prediction}. We also evaluate the effectiveness of the proposed displacement prediction compared with typical classification. The experiment is based on the model with both spatial and temporal modules. 
Tab.~\ref{ablation_loss} shows the displacement prediction with regression loss performs better than classification with cross-entropy loss in the aspect of trajectory selection. Displacement prediction gets rid of hard or manual assignment and converts the classification problem into a regression problem, which helps to converge to better results. It is worth noting that this change will not affect results for $K=6$, demonstrating that trajectory regression and selection are independent tasks.
\begin{table*}[!t]
\begin{center}
   \small
   \centering
   \def\arraystretch{1.2}
   \begin{tabular}{c|c c c|c c c}
      \hline
      {Models} & {minADE$_{1}$} & {minFDE$_{1}$} & {MR$_{1}$} & {minADE$_{6}$} & {minFDE$_{6}$} & {MR$_{6}$}
      \\
      \specialrule{1pt}{0pt}{0pt}
      \hline
      Argoverse Baseline~\cite{chang2019argoverse} & 2.96 & 6.81 & 0.81 & 2.34 & 5.44 & 0.69
      \\
      \hline
      Argoverse Baseline\ (NN)~\cite{chang2019argoverse} & 3.45 &7.88 & 0.87 & 1.71 & 3.29 & 0.54
      \\
      \specialrule{1pt}{0pt}{0pt}
      Jean $(1st)$~\cite{chang2019argoverse, mercat2020multi} & 	1.74 & 4.24 & 0.68& 0.98 &1.42 & \bf 0.13\\ 
      \hline
      uulm-mrm $(2nd)$~\cite{chang2019argoverse} & 1.90& 4.19& 0.63  &0.94 &1.55 &0.22  \\ 
      \hline
      LaneConv~\cite{liang2020learning} & 1.71 & 3.78 & 0.59 & \bf 0.87 & \bf 1.36 & 0.16
      \\
      \hline
      TNT~\cite{zhao2020tnt} & 1.77 & 3.91 & 0.59 & 0.94 & 1.54 & \bf 0.13
      \\
      \specialrule{1pt}{0pt}{0pt}
      \hline
      \def\arraystretch{1.2}
      Ours & \bf 1.66 & \bf 3.69 & \bf 0.588 & \bf 0.87 & 1.38 & 0.158 
      \\
      
      \hline
   \end{tabular}
\end{center}

\caption{The results of our method and top performing approaches on the Argoverse test set.}
\label{test_set_result}
\vspace{-5px}
\end{table*}
 
\begin{table}[htbp]
   \begin{center}
      \small
      \centering
      \def\arraystretch{1.2}
      \begin{tabular}{c|c c}
        \hline
        \multirow{2} * { Metrics} & \multicolumn{2}{c}{Loss} \\
        \cline{2-3} & Classification & Displacement \\
        \specialrule{1pt}{0pt}{0pt}
        {minADE$_{1}$} &1.44 & \bf 1.34\\
        \hline
        {minFDE$_{1}$} & 3.14& \bf 2.95\\
        \hline
        {MR$_{1}$} & 0.54 & \bf 0.50\\
        \specialrule{1pt}{0pt}{0pt}
        {minADE$_{6}$} & 0.74 & \bf 0.73\\
        \hline
        {minFDE$_{6}$} & \bf 1.14 & 1.15\\
        \hline
        {MR$_{6}$} & \bf 0.11 & \bf 0.11 \\
        \hline
        
      \end{tabular}
   \end{center}
   
   \caption{Ablation study on loss design. Classification refers to predict scores and use cross-entropy loss to optimize. Displacement is the displacement prediction with regression loss.}
   \label{ablation_loss}
   \vspace{-5px}
\end{table}


   
\textbf{Data composition}. Since there are mainly three types of data (agent, non-agent vehicles, map) in the Argoverse dataset, we conduct experiments to see whether our TPCN can extract the corresponding features, including map topology relationship and locality. During this experiment, we train our model by removing some specific kinds of data. As shown in Tab.~\ref{ablation_study_data_composition}, we see that our TPCN can model the internal relationship among different types of input data. Map information brings useful topology of the road networks and semantic guidance since most of the driving behavior is based on lane keep and lane changes. Meanwhile, non-agents vehicles provide interaction under some decisions (i.e., nudge, overtake). Lacking any one of the data will lead to a significant performance drop for our TPCN. 

\textbf{Impact of data augmentation}. We conduct an experiment to verify our data augmentation strategy in prediction task. As shown in Tab.~\ref{ablation_study_result}, the data augmentation improves TPCN in terms of all metrics, especially for minFDE.

   

\subsection{Evaluation}
\textbf{Quantitative results}. We compare our model with other methods that achieve the state-of-the-art in Argoverse motion forecasting leaderboard. As shown in Tab.~\ref{test_set_result}, we see that our TPCN improves the metrics for $K=1$ by a large margin and outperforms the existing approaches in $\text{minADE}_{1}$, $\text{minFDE}_{1}$ and $\text{MR}_{1}$ without any complex postprocessing. Moreover, our TPCN is the first method which achieves $\text{minADE}_{1}$ less than 1.7m, $\text{minFDE}_{1}$ less than 3.7m and $\text{MR}_{1}$ less than 0.59. In contrast to existing methods~\cite{chang2019argoverse} that ignore the temporal information or just use 1D CNN or LSTM to encode agents' temporal features, we mutually propagate the spatial and temporal features in order to maintain both locality and temporality. Furthermore, our proposed spatial module can effectively capture better map information compared with LaneConv~\cite{liang2020learning}. Finally, we ranked $1st$, $1st$, $1st$, $2nd$, $3rd$, $5th$ on the leaderboard according to the metrics, respectively.

\textbf{Quantitative results}. We present some multi-modal prediction trajectories on several hard cases shown in Fig.~\ref{fig:prediction_results}. Despite the noise of the input trajectory, TPCN can generate feasible, reasonable, and smooth trajectories with map constraints. For the multi-modality under junction scenarios, TPCN is able to capture the topology of the road network and give possible trajectories along with the lane's successors or neighborhood.

\begin{table}[t]
   \begin{center}
      \small
      \centering
      \def\arraystretch{1.2}
      \begin{tabular}{c|c c c c}
        \hline
        \multirow{2} * { Metrics} & \multicolumn{4}{c}{Data Composition} \\
        \cline{2-5} & none & agents & map & agents + map\\
        \specialrule{1pt}{0pt}{0pt}
        \hline
        {minADE$_{1}$} &2.53 & 1.42 &1.40 & \bf 1.34\\
        \hline
        {minFDE$_{1}$} &3.94 & 3.08 & 3.04& \bf 2.95\\
        \hline
        {MR$_{1}$} &0.81 & 0.55 & 0.54 & \bf 0.50\\
        \specialrule{1pt}{0pt}{0pt}
        {minADE$_{6}$} &1.77 & 0.82 & 0.79 & \bf 0.73\\
        \hline
        {minFDE$_{6}$} &3.54 & 1.32 & 1.29 & \bf 1.15\\
        \hline
        {MR$_{6}$} &0.65 & 0.15 & 0.138 & \bf 0.11 \\
        \hline
        
      \end{tabular}
   \end{center}
   
   \caption{Ablation study of data composition on the Argoverse validation set. Here, agents refer to other agents, and maps refer to map points data.}
   \label{ablation_study_data_composition}
   \vspace{-5px}
\end{table}

\section{Conclusion}
    \vspace{-5pt}
    In this paper, we propose our TPCN that serves as a novel and flexible architecture for prediction learning. TPCN models the motion forecasting problem as joint temporal point cloud learning, consisting of both spatial and temporal modules. In the spatial module, TPCN takes the merit of dual-representation learning in point clouds to maintain better locality and geometric relationships. The temporal module utilizes the proposed Multi-interval Learning and Instance Pooling to capture more fine-grained sequential information. Both modules are learned and propagated mutually to obtain better context information for prediction learning. Experiments on the Argoverse motion forecasting benchmark show the effectiveness of our TPCN.

{\small
\bibliographystyle{ieee_fullname}
\bibliography{egbib}
}

\clearpage
\setcounter{section}{0}
\setcounter{figure}{0}
\setcounter{table}{0}

\begin{center}
    \LARGE{\textit{Supplementary Material}}
\end{center}
\vspace{20pt}

\section{The Detailed Network Architecture}
\label{sec:detailed_net}
\begin{figure}
    \centering
    \includegraphics[width=\linewidth]{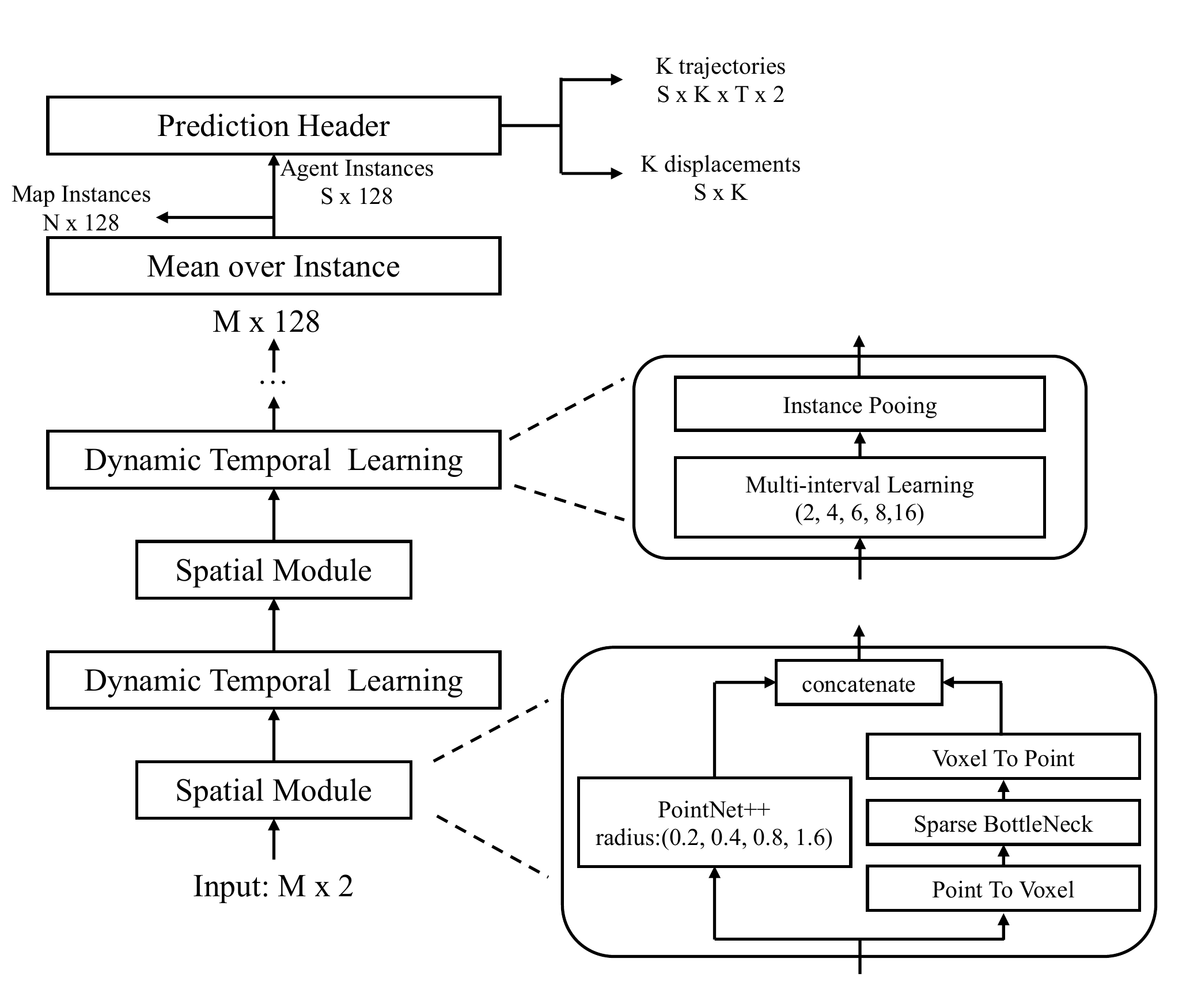}
    \caption{The overall structure of our TPCN. $M$ is the total input points number, including points of $N$ map instances and points of $S$ agent instances. Note each instance has multiple points. $K$ is the number of multi-modal output trajectories. $T$ is the number of prediction timestamps.}
    \label{fig:network}
\end{figure}

We provide the detailed network architecture of our TPCN in Fig.~\ref{fig:network}. The network structure consists of 4 spatial modules and 4 dynamic temporal learning layers. Before the prediction header, we calculate the mean features and remove map instances, which are not required for predictions. Each spatial module takes the output features of the dynamic temporal learning layer as input except the first module. The dynamic temporal learning layer will take the output features of the spatial module as input as well. In the spatial module, the point representation utilizes PointNet++ with neighborhood radius of $[0.2m, 0.4m, 0.8m, 1.6m]$, while the voxel representation uses Sparse BottleNeck as feature extractor. Further, although we use PointNet++ in our point representation learning, we do not apply any sampling or downsampling for the points. We keep all the points in this process.

\section{Ablation Study}
\label{sec:ablation}
\subsection{Interval Selection}
To select the best intervals for Multi-interval Learning, we also do some experiments for the selection of this parameter. As shown in Tab.~\ref{ablation_study_result}, we can see that the performance of our TPCN increases with more intervals. Moreover, the larger interval seems to have a larger impact on our TPCN.
\subsection{Hyperparameters Tuning}
Since we have trajectory regression loss and displacement prediction loss, we need to tune the weight of these two losses. We tried the uncertain loss to learn the weight of different losses. However, the uncertain loss does not improve the performance on the validation set, even leading to a slight decay of the performance. 

As a result, we tried different weights, including $[0.25, 0.5, 0.75, 1]$ and finally choose $1$ as the weight factor. It means that the two losses or tasks should be treated equally.
\begin{table*}[!t]
   \begin{center}
      \small
      \centering
      \def\arraystretch{1.2}
      \begin{tabular}{c |c c c|c c c}
         \hline
         {Intervals}&
         { minADE$_{1}$} & { minFDE$_{1}$} & { MR$_{1}$} & { minADE$_{6}$} & { minFDE$_{6}$} & { MR$_{6}$} \\
         \hline
         $[2]$  & 1.47 & 3.36 & 0.57 & 0.84 & 1.43 & 0.16 \\
         \hline
         $[2, 4]$ & 1.44 & 3.28 & 0.55 & 0.82 & 1.39 & 0.15 \\
         \hline
         $[2, 4, 6]$  & 1.43 & 3.17 & 0.53 & 0.79 & 1.33 & 0.13\\
         \hline
         $[2, 4, 6, 8]$ & 1.39 & 3.07 & 0.51 & 0.76 & 1.24 & 0.12\\ 
         \hline
         $[2, 4, 6, 8, 16]$ & \bf 1.34& \bf 2.95& \bf 0.50 &\bf 0.73 & \bf 1.15 & \bf 0.11 \\
         \hline
      \end{tabular}
   \end{center}
   \caption{Ablation study of intervals on Argoverse validation set.}
   \label{ablation_study_result}
   \vspace{-5px}
\end{table*}

\section{Qualitative Results}
\label{sec:analyzes}

In this section, we provide more qualitative visual results of our TPCN on the Argoverse validation/test set.

\subsection{Qualitative Analysis}
\begin{figure*}[t]
    \vspace{-5px}
    \centering
    \includegraphics[width=1\textwidth]{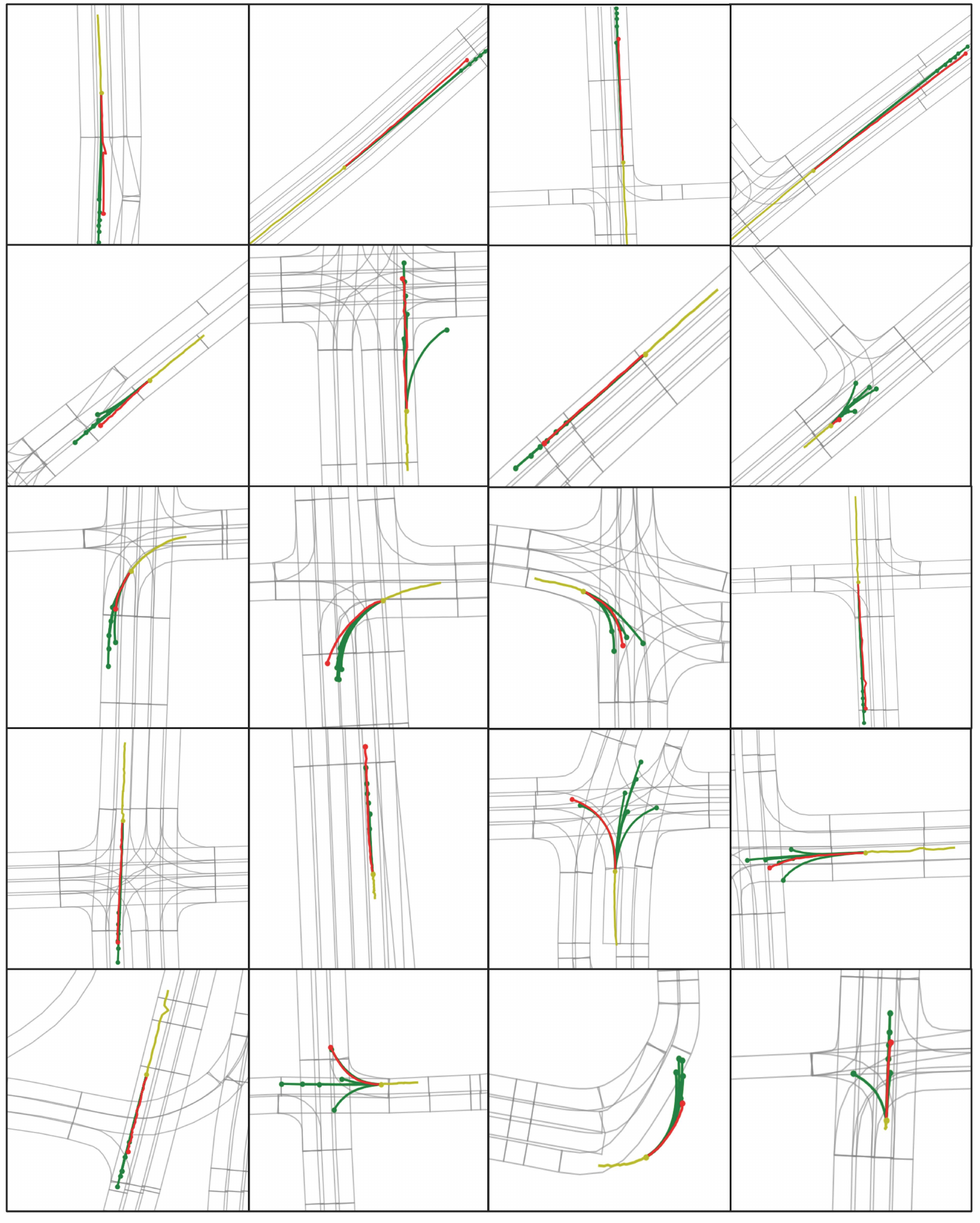}
    \caption{The motion forecasting results on the Argoverse validation set. The target agent's past trajectory is in yellow, predicted trajectory is in green, and ground truth is in red.}
    \label{fig:val_results}
    \vspace{-5px}
\end{figure*}

\begin{figure*}[t]
    \vspace{-5px}
    \centering
    \includegraphics[width=1\textwidth]{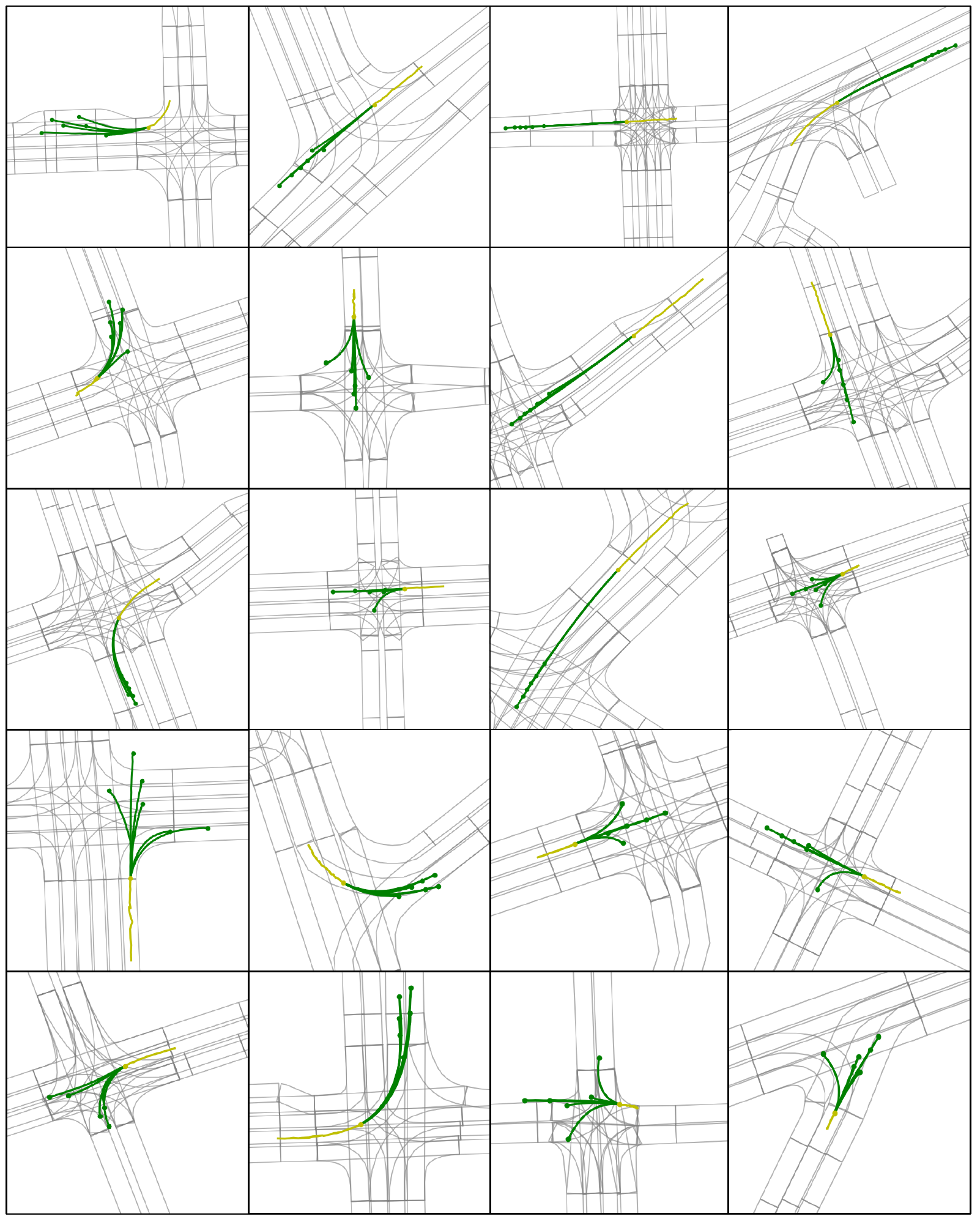}
    \caption{The motion forecasting results on the Argoverse test set. The target agent's past trajectory is in yellow and predicted trajectory in green.}
    \label{fig:test_results}
    \vspace{-5px}
\end{figure*}

We provide more visual results of our TPCN on the Argoverse validation set in Fig.~\ref{fig:val_results}. 
Besides, Fig.~\ref{fig:test_results} shows the results on the Argoverse test set without ground-truth labels.
Generally, these qualitative results demonstrate the effectiveness of our TPCN.

\subsection{Failure Cases}
\begin{figure*}[t]
    \vspace{-5px}
    \centering
    \includegraphics[width=1\textwidth]{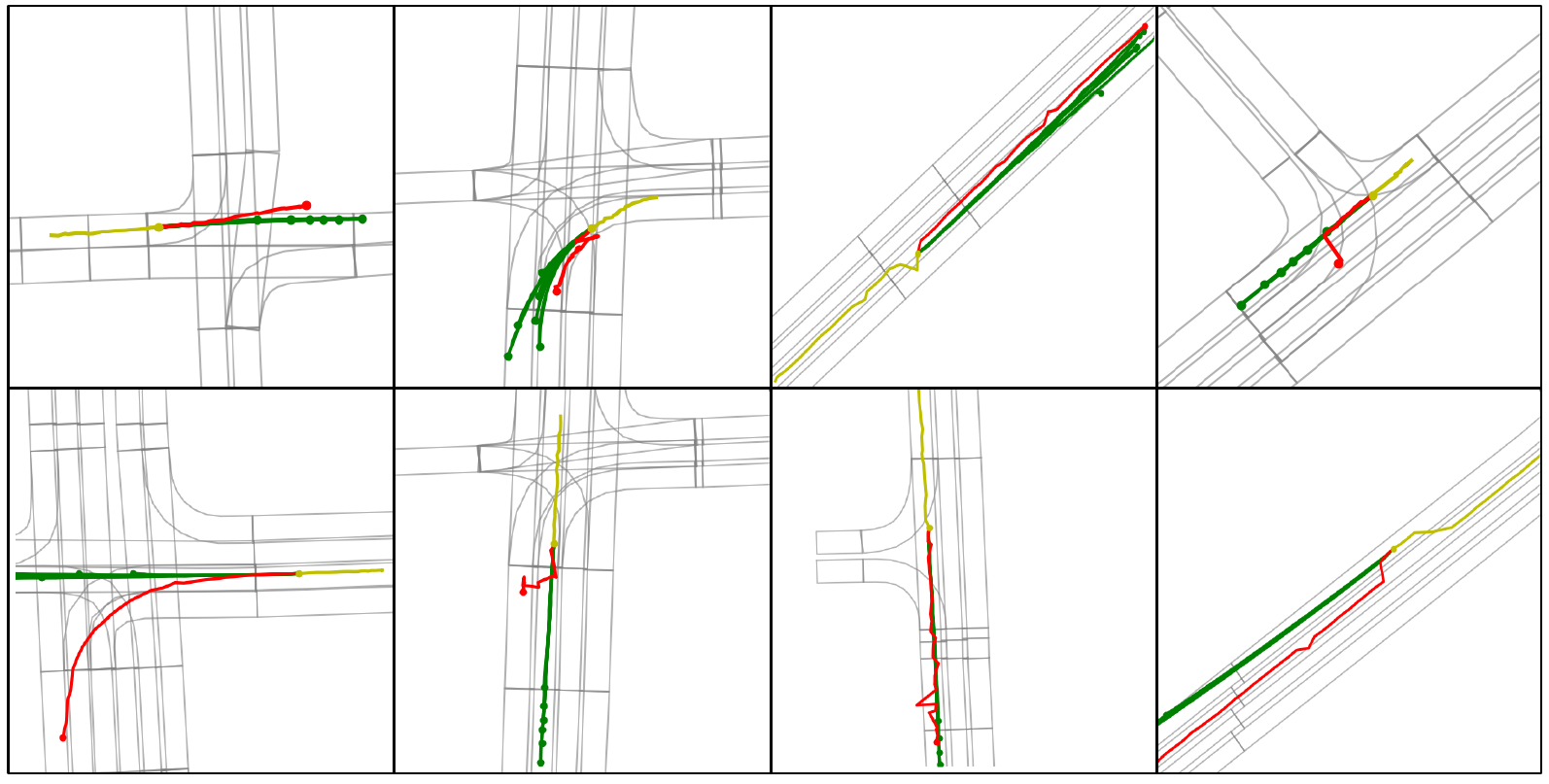}
    \caption{Failure cases on the Argoverse validation set. The target agent's past trajectory is in yellow, predicted trajectory in green and ground truth in red. }
    \label{fig:failure_cases}
    \vspace{-5px}
\end{figure*}

We provide several failure cases on the validation set. As shown in Fig.~\ref{fig:failure_cases}, there are several reasons that cause failure:
\begin{itemize}
    \item The ground-truth labels are not completely correct. Since the data of Argoverse is obtained from tracking, there may be some id switches, leading to the sudden perturbation of the agents' location (e.g., the fourth example in the first row of Fig.~\ref{fig:failure_cases}). Although TPCN does not output a similar trajectory as ground truth, the predicted trajectories are more reasonable and stable without large jerk.
    \item The multi-modality nature of the motion prediction task. There are some cases that the TPCN tends to output the trajectories with more motion constraints rather than map constraints. The first example in the second row of Fig.~\ref{fig:failure_cases} demonstrates this phenomenon. The agent has no prior motion states about left-turn action. Thus TPCN predicts the lane-keeping behavior and ignores the map constraints. Therefore, this can be the further work of our TPCN.
\end{itemize}

\end{document}

%% file: vcl-shortcuts.tex
\usepackage{times}
\usepackage{epsfig}
\usepackage{graphicx}
\usepackage{float}
\usepackage{wrapfig}
\usepackage{amsmath,amssymb,amsthm}
\usepackage{algorithm,algorithmicx,algpseudocode}
\usepackage{bm,xspace}
\usepackage{comment}
\usepackage{verbatim}
\usepackage{multirow}
\usepackage{balance}
\usepackage{url}
\usepackage{booktabs}
\usepackage{etoolbox,siunitx}
\usepackage{calc}
\usepackage{pifont,hologo}
\usepackage[usenames, dvipsnames]{color}
\usepackage{nicefrac}

\usepackage[super]{nth}
\usepackage{nicefrac}
\robustify\bfseries

\DeclareMathSymbol{@}{\mathord}{letters}{"3B}

\def\latex/{\LaTeX}
\def\bibtex/{\hologo{BibTeX}}
